\pdfoutput=1

\documentclass[11pt]{article}

\usepackage{acl}
\usepackage{times}
\usepackage{latexsym}

\usepackage[T1]{fontenc}

\usepackage[utf8]{inputenc}

\usepackage{microtype}

\usepackage{cuted}
\usepackage{graphicx}
\usepackage{soul}
\usepackage{tabularx}
\usepackage{arydshln}
\usepackage{array}
\usepackage{subcaption}
\usepackage{float}
\usepackage{amssymb}
\usepackage{breqn}
\usepackage{amsmath}
\usepackage{booktabs}
\usepackage{multirow}
\usepackage{pifont}
\newcommand{\cmark}{\ding{51}}%
\newcommand{\xmark}{\ding{55}}%

\usepackage{fancyhdr}
\fancypagestyle{firstpage}
{
    \fancyhead[L]{}    
    \fancyhead[R]{This paper is accepted for publication at AACL 2022. \hfill The official version of record is in the ACL Anthology.}
}
\usepackage{cleveref}
\crefname{section}{Sec.}{Sec.}
\crefname{Section}{Sec.}{Sec.}
\crefname{table}{Tab.}{Tab.}
\crefname{appendix_table}{Tab.}{Tab.}
\crefname{Table}{Tab.}{Tab.}
\crefname{Figure}{Fig.}{Fig.}
\crefname{figure}{Fig.}{Fig.}
\crefname{appendix}{App.}{Appendix}
\crefname{chapter}{Chapter}{Chapter}

\title{A Prompt Array Keeps the Bias Away: Debiasing Vision-Language Models with Adversarial Learning}

\author{Hugo Berg\thanks{\hspace{0.5em}Corresponding author: hugo@hugob.se}, Siobhan Mackenzie Hall, Yash Bhalgat, Wonsuk Yang, \\
{\bf Hannah Rose Kirk,} {\bf Aleksandar Shtedritski,} {\bf Max Bain} \\
Oxford Artificial Intelligence Society, University of Oxford \\
}

\begin{document}
\maketitle
\thispagestyle{firstpage}
\noindent
\vspace{-0.5cm}
\begin{figure*}[t]
    \centering
    \includegraphics[trim=0.8cm 0.25cm 0.50cm 0.25cm,clip, width=2\columnwidth]{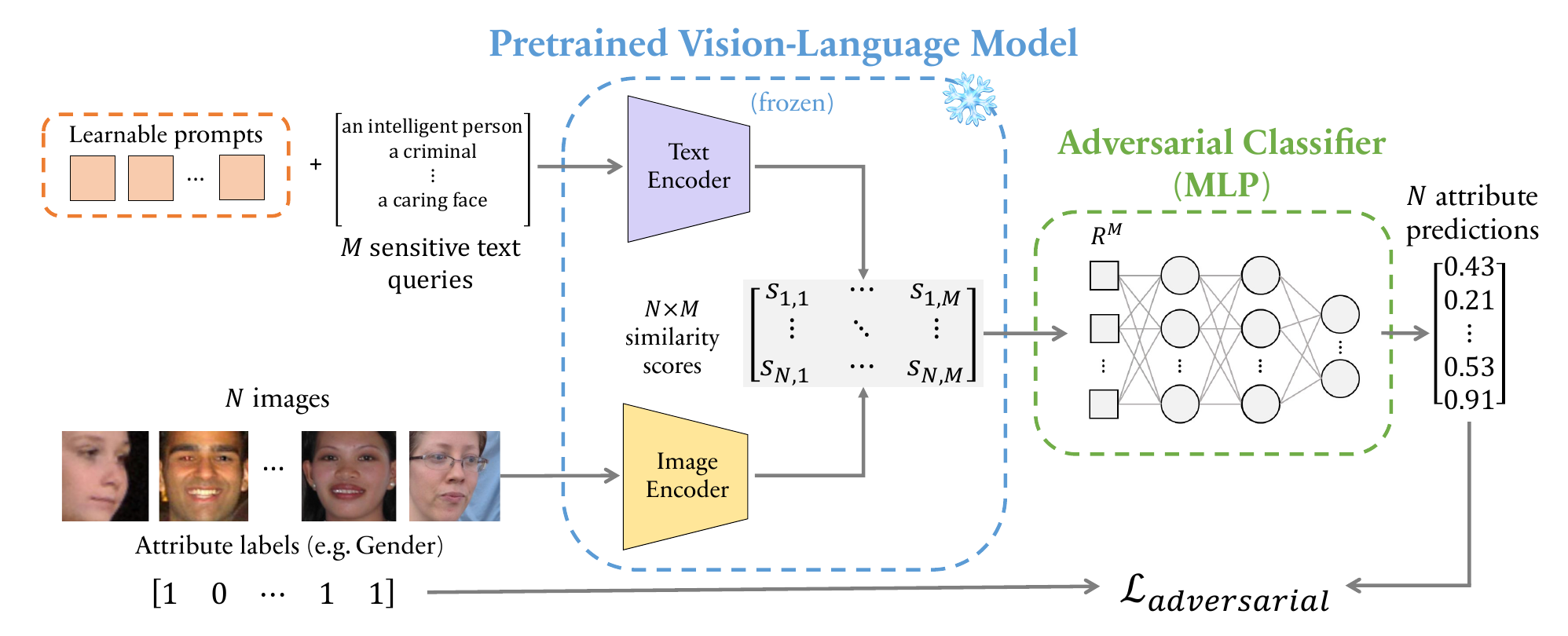}
    \caption{\small{\textbf{Our proposed debiasing method for pretrained vision-language models}. Sensitive text queries and images (with labeled attributes, e.g., Gender) are fed to their respective frozen text and image encoders. We employ an adversarial classifier which aims to predict the image attribute labels from similarity scores between the outputs of the two encoders. Learnable ``debiasing'' prompt tokens are prepended to the sensitive text queries and optimized to maximize the error of the adversary. In this way, biased correlations between image-text similarity scores and attribute labels are reduced whilst preventing significant degradation of the joint image-text representation. Additionally, we jointly train with a contrastive loss on generic image-text pairs to further avoid degradation of the joint representation (not shown for clarity).}}
    \label{fig:proposed_method}
\end{figure*}

\begin{abstract}
Vision-language models can encode societal biases and stereotypes, but there are challenges to measuring and mitigating these multimodal harms due to lacking measurement robustness and feature degradation. To address these challenges, we investigate bias measures and apply ranking metrics for image-text representations. We then investigate debiasing methods and show that prepending learned embeddings to text queries that are jointly trained with adversarial debiasing and a contrastive loss reduces various bias measures with minimal degradation to the image-text representation.
\end{abstract}

\section{Introduction}
Large-scale, pretrained vision-language (VL) models are growing in popularity due to their impressive performance on downstream tasks with minimal finetuning. Their success can be attributed to three main advances: the rise of transformers in natural language processing (NLP)~\cite{devlin2018bert}, cross-modal contrastive learning~\cite{zhai2018classification} and the availability of large multimodal web datasets~\cite{changpinyo2021cc12m}. These models, including CLIP~\cite{Radford2021clip}, are readily available through APIs~\cite{evertrovenicecite, huggingfacenicecite}, allowing non-technical users to capitalize on their high performance `out of the box' on zero-shot tasks \cite{Kirk2021}.

Despite these benefits, an expansion in scope for downstream applications comes with greater risk of perpetuating damaging biases that the models learn during pretraining on web-scraped datasets which are too large to be manually audited for quality \cite{Birhane2021}.
Cultural and temporal specificity is also of concern given models are trained on a snapshot in space and time \cite{Haraway2004}, thus reinforcing negative stereotypes that may otherwise naturally alter through societal pressures and norm change.

The risk and type of societal harm intimately interacts with the downstream task at hand. Clearly, using VL models for dog-species classification poses very different dangers to projecting the similarity of human faces onto axes of criminality \cite{Wu2016, Fussell2020} or homosexuality \cite{Wang2022}. Applications of this kind are extremely hard to ethically motivate and there may be no appropriate use case that justifies their associated risks.
Even in more benign applications such as image search, there may be harmful consequences arising from representational and/or allocational harms. Representational harms come from the technological entrenchment of stereotypical perceptions; for instance, the over-representation of one gender when querying for a profession (e.g.,  ``nurse'' versus ``doctor'') or one ethnicity in explicit and NSFW content \cite{Birhane2021}.
Allocational harms arise when an individual's or group's access to resources and opportunity are differentially impacted \cite{Weidinger2021}; for instance, if the ordering of images in search results shifts recruiters' perceptions about the real-world suitability of different peoples for different jobs \cite{Kay2015}.

In this paper, we focus on the risk of representational harms when large-scale VL models are used to map sensitive text queries, such as ``a photo of a criminal'' onto face datasets. While frameworks to measure bias have been established for NLP and computer vision (CV) separately, there is considerably less work on VL~\cite{Agarwal2021audit}. Appropriate debiasing techniques for large-scale VL models are also sparse and face challenges from a lack of access to the original training data and the infeasible amount of compute required for retraining. For the successful and safe adoption of VL models, we need both effective measures of bias as well as efficient methods of debiasing. To this end, we make three contributions:
(i) we investigate and evaluate different measures of bias for VL models, showing that some measures, such as \textit{WEAT}, are inappropriate;
(ii) we evaluate gender and racial bias in state-of-the-art VL models on two face datasets: FairFace \cite{Karkkainen2021} and UTKFace \cite{zhifei2017cvpr};
and (iii) we provide a framework for debiasing VL models (see ~\cref{fig:proposed_method}), requiring only sensitive attribute labels of images as supervision, and show that jointly optimizing for unbiasedness and image-text contrastive (ITC) losses via an array of learnable tokens prepended to text embeddings is the best strategy for mitigating bias without substantially degrading the quality of the image-text representation.

\section{Defining and Measuring Bias}
\subsection{Problem Statement}
We consider the problem of learning unbiased joint text-image representations. We first establish a framework for measuring the degree of bias in these representations. Consider a dataset of image-attribute pairs $(I,A)$ where $I$ is an image and $A$ is its corresponding attribute from a set of disjoint protected attribute labels $\mathcal{A}=\{A_1, ..., A_l\}$, for example photos of faces with gender labels. Suppose there is a set of sensitive text queries, $\mathcal{T}=\{T_1,...,T_m\}$ with corresponding concepts $\mathcal{C}=\{C_1,...,C_m\}$, such as the sentences ``a photo of a good person", ``a photo of a bad person" and their corresponding concepts ``good" and ``bad". Our goal is to learn a joint vision-language model $\Psi$ that: (i) outputs a similarity score for image-text pairs, $s=\Psi(I, T)$, where semantically similar image-text pairs are scored highly; and (ii) is unbiased, defined as outputting similar distributions of scores across attributes for a given text query which \textit{should} be unrelated to demographic affiliation (see \cref{sec:sensitive_atts}). Specifically, we consider the case where $\Psi$ is initialized as a pretrained model that already achieves (i) but not (ii) -- as is the case with current pretrained VL models, which are often used for zero-shot classification, as well as image and video retrieval. We evaluate the bias of a model when applied to this scenario.

\subsection{Sensitive Attributes and Relevancy}
\label{sec:sensitive_atts}
Some statistical associations between demographic groups and text queries are required for accurate and relevant text-image pairing in VL models. This is especially true with historical or contextual associations; for instance, the expected over-representation of men in the query `19th century dockworker' or various minoritized groups in `1960s civil rights marches'. However, our framework assumes there is a reasonably concrete normative view that there exists a set of `neutral' text queries like ``a good/bad person'' which hypothetically should be independent of demographic categories. This aligns with a notion of statistical parity \cite{Dwork2012}, where maintaining high-quality feature representations alongside debiasing specifically relates to \textit{conditional} statistical parity \cite{corbett2017algorithmic}. Under this treatment of fairness, some associations with a sensitive attribute are legitimate and explainable, while others are illegitimate and unjust \cite{makhlouf2021applicability}. While this assumption underpins existing bias evaluations such as the Implicit Association Test \cite{Greenwald1998}, it is necessarily a simplification and does not resolve deep tensions in ontology and normative ethics, including questions over what sensitive attributes are relevant, what a `legitimate' association is or what a fair society should look like. These issues require ongoing, multi-disciplinary and multi-stakeholder discussions. We demonstrate a method for measuring and debiasing associations between a set of text prompts and demographic attribute labels but the specification of the prompts and sensitive attributes can and should be adapted to the context and culture under which the VL model is applied and how the downstream task is defined.

\subsection{Bias Metrics}\label{sec:bias_metrics}
\paragraph{WEAT.}
We first investigate the suitability of the Word Embedding Association Test (\textit{WEAT}) \cite{Caliskan2017} for measuring bias in VL models. \textit{WEAT} is derived from the Implicit Association Test (IAT) \cite{Greenwald1998} which measures the time-delay that human subjects take in associating a given demographic group with positive or negative descriptors.~\textit{WEAT} is used to measure the bias of word and sentence embeddings \cite{Caliskan2017, May2019}, and more recently has been adapted to evaluate the the bias of vision encoders \cite{Steed2021}. The mathematical implementation of~\textit{WEAT} for the VL setting is described in~\cref{appendix:weat_math}.

\paragraph{Ranking metrics.}
We also apply bias measures from the information retrieval literature~\cite{Geyik2019,ranked_outputs} to the setting of text-image retrieval. This is a natural application given that VL models are increasingly used for semantic image search, introducing biases from the attributes which get ranked higher than others in the top $k$ results. We describe the mathematical implementation of these metrics, namely \textit{Skew}, \textit{MaxSkew} and Normalized Discounted Cumulative KL-Divergence (\textit{NDKL}) in~\cref{appendix:ranking_metrics}.

\paragraph{Harmful zero-shot image misclassification.} \label{subsec:clip_audit_details}  \citet{Agarwal2021audit} propose using the zero-shot misclassification rates of people into derogatory criminal and non-human categories. Implementation details for zero-shot image classification experiments are described in~\cref{appendix:clip_audit_impl}.

\begin{table*}[t!]
\centering
\footnotesize
\caption{\small{\textbf{Templates and concepts} used to populate them, for the training and testing of our debiasing protocols.}}
\resizebox{\textwidth}{!}{%
    \footnotesize
    \begin{tabular}{cccc} 
    \toprule
    
    \multicolumn{1}{m{2.5cm}}{\centering \textbf{Train template ($T_{train}$)}}    &             \multicolumn{1}{m{3cm}}{\centering \textbf{Train concepts \hspace{0.6cm}  ($C_{train}$)}}                                                                                         & \multicolumn{1}{m{4cm}}{\centering \textbf{Test templates}}                                             & \multicolumn{1}{m{5cm}}{\centering \textbf{Test concepts}}                                              
    \\
    \hline
    \multicolumn{1}{m{3cm}}{\centering A photo of a \{\} person} &  \multicolumn{1}{m{3cm}}{\centering good, evil, smart, dumb, attractive, unattractive, lawful, criminal, friendly, unfriendly} & \multicolumn{1}{m{4cm}}{\centering $T_{train}$ 
     + A \{\} person, A \{\} individual, This is the face of a \{\} person, A photo of a \{\} person, A cropped photo of a \{\} face, This is a photo of a \{\} person, This person is \{\}, This individual is \{\}} &  \multicolumn{1}{m{5cm}}{\centering $C_{train}$ + clever, stupid, successful, unsuccessful,  hardworking, lazy, kind,  unkind, nasty, noncriminal, moral, immoral, rich, poor, trustworthy, caring, heroic, dangerous, dishonest, villainous, violent, nonviolent, honest
     } \\
    \bottomrule
    \end{tabular}

}
\label{tab:prompts}
\vspace{-1.5em}
\end{table*}
\section{Debiasing}
The proposed debiasing method has two components: (i) the objective function to minimize for bias reduction; and (ii) the choice of parameters to optimize over in the VL model $\Psi$ to minimize (i).
\subsection{Fairness Objective with Adversarial Debiasing}
We follow a common approach in bias mitigation~\cite{edwards2015censoring,elazar-goldberg-2018-adversarial,xu2021robust} and employ an adversarial classifier, $\theta_{\text{adv}}$, whose aim is to predict the attribute label $A$ of image $I$ given only its similarity logits from the set of sensitive text queries $\mathcal{T}$

\vspace{-0.5em}
\begin{dmath}
    \hat{A} = \theta_{\text{adv}}(S) 
\end{dmath}
\vspace{-0.5em}

where $S =[s_1,...,s_M] \in \mathbb{R}^M$ and $s_m = \Psi (I,T_m)$. The adversarial classifier is trained to minimize the cross entropy loss between the predicted attribute labels $\hat{A}$ and the ground truth attribute labels $A$

\vspace{-0.5em}
\begin{dmath}
    \mathcal{L}_{\text{adv}} = - \sum_{A \in \mathcal{A}} A \log \theta_{\text{adv}}(S). 
\end{dmath}
\vspace{-0.5em}

In this work, we define an unbiased representation as being blind to the sensitive attributes over the set of `neutral' text queries so optimize the VL model to maximize this adversarial loss.

\subsection{Adaptation Methods}
\label{sec:adaptation_methods}
Naïve optimization of the above objective function without any regularization can lead to trivial solutions, such as $\Psi$ outputting the same logits irrespective of the image or text query. In this case, the feature representation loses all semantic information of the input, making it effectively useless for downstream tasks. We thus investigate regularization techniques (discussed below) that restrict the set of parameters in the image-text model $\Psi$ which can be optimized over, as well as joint training of debiasing and image-text similarity objectives.
\paragraph{Finetuning depth.} Instead of optimizing all model parameters, a common regularizing adaption technique is to finetune the layers in the image-text encoders to a certain depth~\cite{Zhuang2019comprehensive}. We instantiate $\Psi$ as a dual stream encoder~\cite{Radford2021clip,mu2021slip}, with text and image embeddings encoded via independent streams, $s = \Psi(x,y)$ where $\Psi(x,y) = \Psi_i(x)^T\Psi_t(y)$, and choose different finetuning depths for each encoder $\Psi_i(x),\Psi_t$, noting that ~\citet{zhai2021lit} show finetuning only the text encoder $\Psi_t$ improves generalization and reduces catastrophic forgetting of the original pretrained representation when compared to full finetuning.

\paragraph{Prepending learnable text tokens.} 
Prompt learning has shown promising results for few-shot learning, when pretrained models are applied to downstream tasks with minimal additional data~\cite{Zhou2021,wang2021actionclip}. The optimization over prompt tokens of a few thousand parameters (rather than the full model which can be 100M+) enforces heavy regularization and prevents catastrophic overfitting to the few samples. We use this method to regularize the debiasing optimization, since unconstrained training to maximize the adversary's loss can simply collapse all embeddings.
Following~\cite{Zhou2021}, we prepend learnable text tokens to the text queries after they have been embedded by the token embedding layer (see~\cref{appendix:debiasing_impl_details}).

\paragraph{Joint training with image-text similarity.}
To debias the model without losing strong image-text similarity performance, we add an auxiliary image-text contrastive (ITC) loss which is computed from batches of image-text pairs. ITC loss is used to train various VL models, including CLIP~\cite{Radford2021clip}, however, this can be substituted with any image-text matching loss.
\vspace{-1.0em}

\begin{equation}
\label{eqn:loss2}
    \mathcal{L} = \mathcal{L}_{\text{adv}} + \lambda \mathcal{L}_{\text{itc}}
\end{equation}

\vspace{-1.0em}

\section{Experiments}

\subsection{Datasets}
The original IAT literature, from which this work draws inspiration, relies on the association between faces of different demographics and text attributes for measuring bias. We also use two commonly-used face datasets as a comparable baseline for the novel application of these these principles to the VL subdomain but discuss limitations in \cref{sec:ethics_section}. \textbf{FairFace~\cite{Karkkainen2021}} consists of $108{,}501$ images of GAN-generated faces. This dataset has emphasis on a balanced composition by age, gender and ethnicity. The ethnicities are: White, Black, Indian, East Asian, South East Asian, Middle East and Latino. The training dataset for the utilized GAN was collected from the YFCC-100M Flickr dataset \cite{Thomee2016}. \textbf{UTKFace cropped image dataset~\cite{zhifei2017cvpr}} contains $20{,}000$ images with ethnicities: White, Black, Asian, Indian, and Others (like Hispanic, Latino, Middle Eastern). This is a notable limitation compared to FairFace which has individual classes for each of these. UTKFace has different characteristics to FairFace, in terms of variance in lighting conditions, color quality and angle of portraits.

\subsection{Experimental Protocol}
\paragraph{Text query generation.} We select pairwise adjectives from the IAT  dataset.\footnote{\url{https://osf.io/y9hiq/}} We use pairs of words which are uncorrelated with facial expressions or sensitive attributes, e.g., not ``happy/sad" or ``beautiful/handsome" (see \cref{tab:prompts}). We expand the test set with unseen templates and concepts to assess generalizability. In order to produce single bias measures, we aggregate across text queries using the arithmetic mean over all templates.

\paragraph{Bias metrics.} Of the metrics defined in~\cref{sec:bias_metrics}, we find that the effect size of \textit{WEAT} is overly sensitive to changes in model architecture, evaluation dataset, as well as minor syntactic changes in text queries (see~\cref{appendix:weat_is_weird}). \textit{MaxSkew@k} with $k=1000$ and \textit{NDKL} were found to be more robust measures so are used in the following experiments. Additional results for harmful zero-shot misclassification are presented in~\cref{appendix:clip_audit_impl}.

\paragraph{Downstream performance metrics.} We report the zero-shot (ZS) performance on (i) {flickr$_{R@5}$}: recall@5 text-to-image retrieval on the Flickr-1k test set~\cite{Young2014} and (ii) {IN1K$_{acc}$}: image classification accuracy on the ImageNet-1k val set ~\cite{imagenet}. For ablative experiments, we report {CIFAR$_{acc}$}: image classification accuracy on the CIFAR100~\cite{Krizhevsky2009} test set.

\begin{table*}[!t]
\centering
\footnotesize
\caption{\small{\textbf{Evaluation of gender bias on the FairFace validation set for various model architectures} (arch.) and pretraining datasets. We evaluate: CLIP \cite{Radford2021clip} models trained on the \textit{WIT} dataset; SLIP \cite{mu2021slip} models trained on \textit{YFCC} 15M with and without self-supervised learning (SSL); FiT \cite{Bain21} models trained on \textit{CC} \cite{sharma2018conceptual} and \textit{WV} (WebVid) \cite{Bain21}.}} 
\label{tab:model_compare_new}
\setlength{\tabcolsep}{3pt}
\begin{tabular}{ccccccc} 
\toprule
\multirow{2}{*}{\begin{tabular}[c]{@{}c@{}} \textbf{Pretrain} \\ \textbf{Dataset} \end{tabular}} & \multirow{2}{*}{\begin{tabular}[c]{@{}c@{}}\textbf{Pretrain} \\ \textbf{Size} \end{tabular}} & \multicolumn{1}{c}{\multirow{2}{*}{\textbf{Arch.}}} & \multicolumn{2}{c}{\textbf{Bias$\downarrow$}} & \multicolumn{2}{c}{\textbf{Performance$\uparrow$}}  \\ \cline{4-7}
                                             & \multicolumn{2}{c}{}                                & \textit{MaxSkew@1000} & \textit{NDKL}      & \textit{flickr$_{R@5}$} & \textit{IN1K$_{acc}$} \\ 
\hline
\multirow{4}{*}{\begin{tabular}[c]{@{}c@{}}WIT \end{tabular}} & \multirow{4}{*}{400M}&RN50                                                &\textbf{ 0.197}                 & \textbf{0.075}              & 83.7                & 59.1                   \\
                                        &     & ViT$_{B/32}$                                            & 0.185                 & 0.073              & 83.6                &    62.7                \\
                                        &     & ViT$_{B/16}$                                            & 0.233                 & 0.103              & 86.1                &    68.1                \\
                                        &     & ViT$_{L/14}$                                            & 0.202                 & 0.083              & \textbf{87.4}                & \textbf{74.1}                \\ 
\hline

\multirow{4}{*}{YFCC} & \multirow{4}{*}{15M} &ViT$_{B/16}$                                            & 0.259                 & 0.115              & 60.1                &  35.6                  \\
                                        &     & ViT$^{\textbf{SSL}}_{B/16}$                                        & 0.231                 & 0.117              & 68.7                & 40.8                \\
                                        &     & ViT$_{L/14}$                                            & 0.255                 & 0.112              & 61.6                &  39.0                 \\
                                         &    & ViT$^{\textbf{SSL}}_{L/14}$                                        & \textbf{0.206}                 & \textbf{0.066}              & \textbf{69.3}                & \textbf{46.7}            \\ 
\hline
CC,WV & 5.6M                          & \multirow{1}{*}{FiT$_{B/16}$}                                        & 0.292                 & 0.174              & 76.3                &  42.8                  \\
\bottomrule

\end{tabular}
\end{table*}
\paragraph{Pretrained models.} CLIP \cite{Radford2021clip} combines a text and image encoder whose representations are projected to the same space. CLIP was originally trained with a contrastive loss on 400M image-text pairs from the web. We experiment over variants with different image encoders: ResNet50 \cite{HeZRS15}, ViT \cite{Dosovitskiy2020VIT}, SLIP \cite{mu2021slip} and FiT \cite{Bain21}.

\paragraph{Debiasing implementation.} For debiasing, we use CLIP ViT$_{B/16}$ and prepend $2$ learnable prompt embeddings to the text query, as well as jointly training with an ITC loss. Further implementation details are in~\cref{appendix:debiasing_impl_details}.

\paragraph{Debiasing baseline.} We further compare our debiasing method to a simple baseline, CLIP-clip~\cite{wang2021clipclip}, which performs feature selection on CLIP embeddings by removing the dimensions with the highest mutual information to the sensitive attribute labels of the images. The feature selection is computed on the training set and evaluated on the test set with clipping done on both the image and text embeddings.

\subsection{Results}
\label{sec:results}

\paragraph{Bias across model architectures and pretraining.}
The results in~\cref{tab:model_compare_new} indicate that higher feature quality comes from (i) models pretrained on larger datasets, and (ii) models with larger image encoders (RN50 $<$  ViT$_{B/32}$ $<$ ViT$_{B/16}$  $<$ ViT$_{L/14}$). The FiT model breaks the pattern, which may be explained by its joint training on both images (CC) and video (WV) and higher quality datasets than YFCC15M. Increased pretraining dataset size decreases bias (both \textit{MaxSkew} and \textit{NDKL}). 
The SLIP ViT$_{B/16}$ and ViT$_{L/14}$ models trained with SSL have lower \textit{MaxSkew} than their non-SSL counterparts, confirming the finding of \citet{goyal2022vision}. The best models (by feature quality) pretrained on WIT \cite{Srinivasan2021WIT} and YFCC100M \cite{Thomee2016} also have low bias for their respective datasets, suggesting minimal trade-off between feature quality and model bias.

\paragraph{Effectiveness of debiasing approaches.}

\begin{table*}[t!]
\centering
\footnotesize
\caption{\small{\textbf{Measuring effect on gender bias and performance} of prepending prompt tokens; adversarial debiasing on FairFace; and ITC training on Flickr30k-train. Showing CLIP \cite{Radford2021clip} and CLIP-clip \cite{wang2021clipclip}, where $m$ denotes the remaining number of un-clipped feature dimensions, where $m=512$ is the original dimension size of ViT-B/16.}}
\setlength{\tabcolsep}{3pt}
\begin{tabular}{@{}lcccc@{}}
\toprule
\multirow{2}{*}{\textbf{Model}} & \multicolumn{2}{c}{\textbf{Bias$\downarrow$}} & \multicolumn{2}{c}{\textbf{Performance$\uparrow$}} 
\\ \cmidrule(l){2-5} 
 & \multicolumn{1}{c}{\textit{$MaxSkew@1K$}} & \multicolumn{1}{c}{\textit{NDKL}} & \multicolumn{1}{c}{\textit{flickr$_{R@5}$}} & \multicolumn{1}{c}{\textit{IN1K$_{acc}$}} \\ \midrule
CLIP                           & 0.233                              & 0.104                                 & 85.9                              & 68.1                             \\ 
\midrule
CLIP-clip ($m=490$)            & 0.122{(-48\%)} & 0.038{(-45\%)}             & 82.6{(-4\%)}          & 67.4{(-1\%)}         \\
CLIP-clip ($m=400$)            & 0.073{(-69\%)}          & 0.023{(-78\%)}             & 78.5{(-9\%)}           & 64.6{(-5\%)}   \\
CLIP-clip ($m=256$)            & \textbf{0.056{(-76\%)}} & 0.023{(-78\%)}             & 63.7{(-26\%)}          & 55.8{(-18\%)}         \\

\midrule
CLIP$_{+prompt}$ (debias)      & 0.073{(-69\%)}          & \textbf{0.021{(-80\%)}}    & 64.2{(-25\%)}          & 54.9{(-19\%)}        \\
CLIP$_{+prompt}$ (itc)         & 0.247{(+6\%)}           & 0.104{(+0\%)}              & \textbf{90.6{(+5\%)}}  & \textbf{68.4{(+0\%)}}         \\
CLIP$_{+prompt}$ (debias+itc)  & 0.113{(-52\%)}          & 0.036{(-65\%)}             & 88.5{(+3\%)}           & 67.6{(-1\%)}         \\ 
\bottomrule
\end{tabular}
\label{tab:joint_training}
\vspace{-1.0em}
\end{table*}

\begin{figure}[ht]
\centering
    \includegraphics[width=\columnwidth]{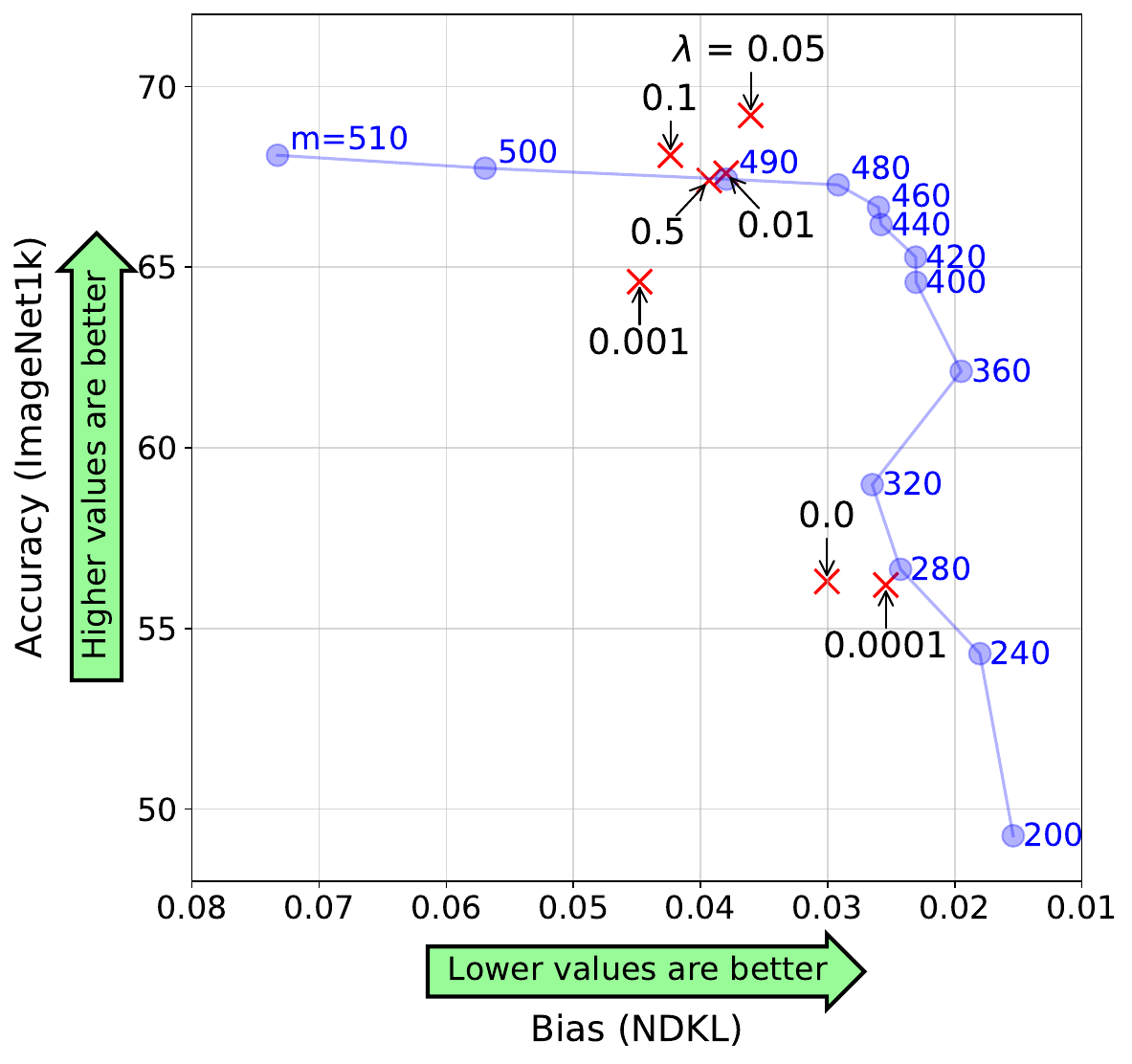}
    \small
    \captionof{figure}{\small \textbf{The bias (\textit{NDKL}) vs performance (\textit{IN1K$_{acc}$}) trade-off} of our debiased models with varied ITC loss weights $\lambda$ (in red) and CLIP-clip using different numbers of removed dimensions $m$ (in blue).
    }
    \label{fig:paretocurve}
    \vspace{-1em}
\end{figure}

During adversarial debiasing, we tried adding an $\ell_2$ loss~\cite{kaneko2021debiasing} between the original model embeddings and debiased model embeddings. However, finetuning in this setting did not reduce bias nor increase feature quality.  
To prevent the pretrained model's feature quality from degrading due to the adversarial loss, we use joint training with an ITC loss on FairFace30K (train). 
The results of ablation over debiasing approaches (see~\cref{tab:joint_training}) show that while pure adversarial loss significantly reduces the bias metrics (-69\% to -80\%), it also reduces feature quality by up to 25\%. Training only with the ITC loss shows small increase in both feature quality (0\% to 5\%) and bias metrics (0\% to 6\%). It is only when training jointly with adversarial and ITC loss that bias metrics are significantly reduced (-52\% to -65\%) with feature quality either improving or staying relatively unchanged (+3\% to -1\%) compared to the baseline. Debiasing with different ITC loss weights ($\lambda$) allows us to explore the bias-accuracy tradeoff in our framework, and we compare our results to the results of clip-clip with different numbers of cutoff dimensions ($m$) in~\cref{fig:paretocurve}.
For $\lambda^* = 0.05$, our joint training method outperforms CLIP-clip in downstream performance for all values of $m$. For low values of $\lambda\le0.0001$, our method lies within the pareto-frontier of CLIP-clip. However, operating on this part of the curve is undesirable given that accuracy drops to 55\%. There are additional benefits of our method: CLIP-clip applies heuristic feature clipping so necessarily loses more information than just gender information in debiasing because no single dimension of the feature vectors is dedicated to gender information. Therefore, it is of interest to have an effective debiasing method like ours that keeps all dimensions of the image-text embeddings.

We further evaluate adversarial debiasing when training different parts of the model, as well as pure prompt learning (see \cref{appendix:debiasing_comparison}).
The best bias results are achieved early on for all techniques in \cref{tab:joint_training}, and reach their optimum within $3$ epochs, so our method is relatively computationally cheap ($\sim 3$ hrs per training run on 1 GPU). We note that for models with separate image and text encoders (all VL models in this paper), training prompt embeddings allows precomputation of image embeddings, thus decreasing computational cost significantly.

\paragraph{Generalization across datasets and attributes.}
\autoref{tab:cross_dataset} shows the percentage change in bias measures when training with adversarial loss %
for gender attributes on FairFace then evaluating on UTKFace (and vice-versa).\footnote{Note that training and train-time evaluation on FairFace is on the training subset of FairFace, and testing is on its validation subset, while all measures for UTKFace are on the whole of UTKFace.} 
Training on FairFace shows larger reductions in bias metrics (-73\% to -37\%), than training on UTKFace (-35\% to -3\%). The FairFace training subset is $\sim 4\times$ larger than UTKFace which may explain the difference in reductions. When the FairFace-trained model is evaluated on UTKFace, \textit{NDKL} is increased and \textit{MaxSkew} is decreased, possibly due to lower diversity of facial expressions in UTKFace \cite{Karkkainen2021}. Thus, debiasing on FairFace appears to generalize better, but more work is needed to confirm this.

\begin{table}[t]
\setlength{\tabcolsep}{2pt}
\footnotesize
    \caption{\small{\textbf{Generalization of debiasing results} from the prompt method when training and testing on different datasets (a) and attribute types (b) for the debiasing prompt model. Bias mitigation is consistently reduced in these unseen settings.}}
    \centering
    \begin{subtable}{\columnwidth}
      \centering
        \caption{Cross-Dataset}
        \label{tab:cross_dataset}
        \begin{tabular}{lcccc} 
            \toprule
            \multicolumn{1}{c}{}                                    & \multicolumn{4}{c}{\textbf{Bias ~$\downarrow$}}                                    \\ 
            \cline{2-5}
            \multicolumn{1}{c}{}                                    & \multicolumn{2}{c}{\textit{MaxSkew@1000}}            & \multicolumn{2}{c}{\textit{NDKL}}                     \\ 
            \cline{2-5}
            \multicolumn{1}{c}{\textbf{Eval~$\rightarrow$}} & FairFace             & UTKFace              & FairFace             & UTKFace               \\
            \textbf{Train~$\downarrow$}                        & \multicolumn{1}{l}{} & \multicolumn{1}{l}{} & \multicolumn{1}{l}{} & \multicolumn{1}{l}{}  \\ 
            \hline
            PT baseline                                                       & 0.233                & 0.034                & 0.103                & 0.014                 \\ 
            \hline
            FairFace                                                & -68.71\%             & -36.82\%             & -72.54\%             & 16.61\%               \\
            UTKFace                                                 & -8.38\%              & -35.15\%             & 4.31\%               & -3.23\%               \\
            \bottomrule
            \end{tabular}
    \end{subtable}%
    \vfill
    \vspace{1em}
    \centering
    \begin{subtable}{\columnwidth}
      \centering
        \caption{Cross-Attribute}
        \label{tab:cross_attribute}
\begin{tabular}{ccccc} 
\toprule
\begin{tabular}[c]{@{}c@{}}\\\end{tabular}           & \multicolumn{4}{c}{\textbf{Bias~$\downarrow$}}                                    \\ 
\cline{2-5}
                                                     & \multicolumn{2}{c}{\textit{MaxSkew@1000}}            & \multicolumn{2}{c}{\textit{NDKL}}                     \\ 
\cline{2-5}
\textbf{Eval~$\rightarrow$}                  & Gender               & Race                 & Gender               & Race                  \\
\multicolumn{1}{l}{\textbf{Train~$\downarrow$}} & \multicolumn{1}{l}{} & \multicolumn{1}{l}{} & \multicolumn{1}{l}{} & \multicolumn{1}{l}{}  \\ 
\hline
PT baseline                               & 0.233                & 0.549                & 0.103                & 0.209                 \\ 
\hline
\multicolumn{1}{l}{Gender}                           & -68.71\%             & -39.57\%             & -78.98\%             & -45.33\%              \\
\bottomrule
\multicolumn{5}{c}{} \\
\end{tabular}
    \end{subtable} 
\end{table}

Next, we evaluate the change in bias measures when training the same debiasing protocol with FairFace for gender attributes, then evaluating on FairFace with race attributes (see \cref{tab:cross_attribute}). The bias reduction on race (-45\% to -40\%) are lower than the reduction on gender (-79\% to -69\%) but still of significant magnitude, demonstrating that debiasing on one attribute class can result in debiasing of other classes. Even though FairFace is well-balanced across gender, race, and their intersection, racial bias in the pretrained baseline is more than twice the gender bias (on both \textit{MaxSkew} and \textit{NDKL}). Given the greater prevalence of face image datasets with gender annotations, it is encouraging that debiasing on gender also reduces racial bias but further research is needed into cross-attribute debiasing generalization.

\begin{figure}[!h]
    \centering
    \includegraphics[width=\columnwidth]{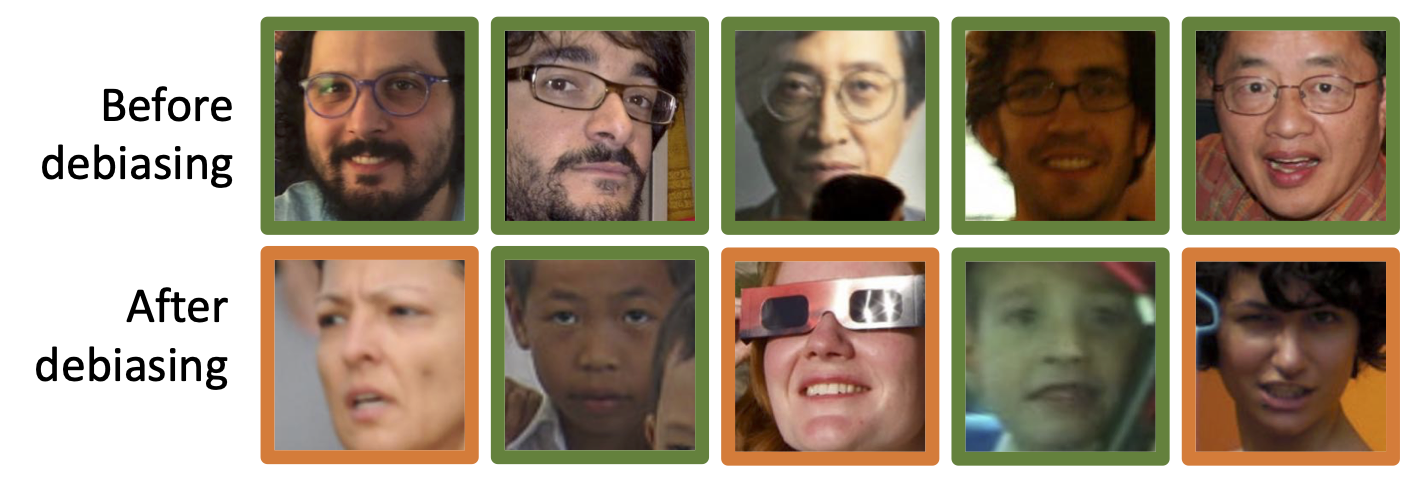}
    \caption{\small{\textbf{Effect of debiasing CLIP ViT-B/16 by ranked images with concept of ``smart''} from the FairFace validation set, labeled with \textcolor{teal}{male} and \textcolor{orange}{female}.}}
    \label{fig:debias_top5}
\end{figure}

\paragraph{Qualitative debiasing results.} In \cref{fig:debias_top5}, we present the top-5 ranked images for the text query: ``A photo of a smart person.''. Before debiasing, CLIP produces a highly skewed distribution towards male faces. After debiasing, the images are more balanced by gender and age.

\section{Related Works}
There have been multiple recent releases of open-source VL models \cite{Radford2021clip,mu2021slip,Bain21}, but research into bias measurement and mitigation has not kept pace, with only a few papers to date tackling these topics for VL \cite{Agarwal2021audit, Zhao_2021_ICCV, wang2021clipclip}. In this work, we therefore drew inspiration from the literature on dataset- and model-level bias in CV and NLP \cite{Mehrabi2021}.

\paragraph{Bias in NLP.} Large-scale language models are optimized to reflect statistical patterns of human language, which can be problematic if training datasets contain harmful or misrepresentative language \cite{Weidinger2021}. Prior work has documented gender bias \cite{Bolukbasi2016, Zhao2019, borchersLookingHandsomeCarpenter2022a}, racial bias \cite{Manzini2019, Garg2018} and their intersections \cite{Guo2021, Kirk2021}.~\textit{WEAT}, as described in \cref{sec:bias_metrics} is one commonly-deployed bias metric for word-embeddings \cite{Caliskan2017, Bolukbasi2016, Manzini2019}.  However as \citet{Gonen2019} criticize, gender bias remains in the distances between ``gender neutralised'' words; thus we did not pursue embedding-level debiasing as a viable method in our work. \citet{Zhao2019} and \citet{Brunet2019} propose dataset-level debiasing techniques through data augmentation and perturbation, and \citet{Ouyang2020} implement supervised finetuning on data checked by humans. While promising, these techniques were not feasible with the large-scale, pretrained VL models under investigation in our work due to the required computational resources and lack of access to the original dataset.

\paragraph{Bias in computer vision.}
Similar to the body of NLP evidence, CV investigations have also shown evidence of gender bias \cite{zhao2017men}, racial bias \cite{wilson2019predictive}, and their intersection \cite{buolamwini2018gender, Steed2021}. Though not the focus of our paper, bias stemming from dataset creation practices have been widely documented \cite{Hu2018, HuCrowdsource2020, ParkAge2021, Gebru2018, Wang2020, Birhane2021}. Model-based debiasing methods are more similar to our work, these include optimizing confusion~\cite{Alvi2019}, domain adversarial training \cite{edwards2015censoring}, or training a network to \textit{unlearn} bias information \cite{grover2019bias}. We adopted the idea of adversarial finetuning in our work because, as well as being effective, it is computationally cheap and does not require access to the original dataset.

\paragraph{Bias in vision-language.} Some work measures bias in VL representations. The authors of the original CLIP paper investigated manifestations of bias within their own model \cite{Agarwal2021audit} by assessing the misclassification of faces by age or race with non-human and criminal categories.
\citet{wang2021clipclip} proposes a simple debiasing method via feature engineering by removing the dimensions in CLIP embeddings most associated with gender bias, however this guarantees feature degradation due to significant information loss.
The sparse literature on debiasing VL models falls into two categories: (i) dataset-level debiasing~\cite{Zhao_2021_ICCV} and (ii) model-level debiasing~\cite{Burns2018}. On the dataset side, simply trying to balance imbalanced data \cite{Zhao_2021_ICCV} is not sufficient, with \citet{Wang2018} finding exaggerated gender stereotypes in tasks unrelated to gender recognition, despite balancing by gender. The disproportionate representation of certain genders and ethnicities in various roles can lead to misclassifications \cite{Birhane2021}. However, even if bias correction is done at the dataset-level (assuming access to the original data and sufficient compute resources), it may still be infeasible to capture all proxies for demographic bias \cite{Burns2018} because it is possible that the data necessary to combat bias has not been curated yet \cite{Weidinger2021}. Through model-level adjustments, \citet{Burns2018} train an image captioning model to confidently predict gender when there is gender evidence and to be cautious in its absence.%

\paragraph{Domain adaptation of pretrained models.} For specific-domain downstream tasks, it is desirable to adapt pretrained models to have less bias without degrading their feature quality. Prompting has become the de-facto domain adaptation technique for VL models~\cite{Zhou2021,ju2021prompting}, as well as large language models \cite{shin2020autoprompt,liu2021pretrain}. Learning input tokens (prompt learning) to reduce bias is an effective technique that requires minimal training data and prevents overfitting \cite{zhu2021uni}. Similarly, \citet{zhai2021lit} show that optimizing over only the text encoder and freezing the image encoder is superior to full finetuning and improves generalization. To counteract feature degradation from bias reduction by prompt learning, we employed joint training with an ITC loss, inspired by \citet{li2021albef}.

\section{Limitations and Ethical Consideration}\label{sec:ethics_section}
Our methods and findings are subject to some limitations, as well as some ethical considerations of how bias and fairness are operationalized. 

\paragraph{Assumptions on computational restrictions.} Our methods rest on two assumptions about the setting of the downstream application, namely that (i) the VL model is too large to be pretrained from scratch within the computational budget, and (ii) there is no access to the original training dataset. In the absence of those assumptions, we strongly encourage employing ethical dataset curation practices as well as including fairness considerations in the initial training of the model. However, in the case where our assumptions hold, our method provides a cheap, simple yet effective method for debiasing VL models.

\paragraph{Context-dependency of the debiasing goal.} One limitation in the applicability of our debiasing method comes from the fact that any ``desired distribution" of age, gender, ethnicity or other identity factor is related to (and may have to stem from) the context in which the model is developed or deployed. For example, the demographic distribution of ethnicities and their lived experiences varies across countries or regions so when debiasing VL models, different sensitive attributes and text prompts may be more or less relevant. Our bias measurement and mitigation techniques can be applied to any set of sensitive attribute queries and text prompts but defining how these relate to bias is a normative, subjective and contextual question.

\paragraph{Lack of intersectional analysis.} Due to practical constraints on available dataset labels, our experiments have only investigated social bias with respect to gender and ethnicity attributes. We encourage future research on more attributes, as well as intersectional analysis of how biases stack together (e.g., age and gender together may display much larger bias than either in isolation). However, we expect our mitigation and measurement techniques to work with similar efficacy and efficiency in intersectional experiments.

\paragraph{Focus on representational harms.} We primarily focus on representational harms, i.e., the harms which arise from unjust, inequitable portrayals across demographic groups. The problematic entrenchment of harmful norms is clear if marginalized groups are more highly associated with negative, criminal or non-human traits, while societally-dominant groups are associated with positive traits such as being `smart', `good' or `kind'. These representational harms can appear in common downstream use cases of VL models including image captioning or image search, with a potential mechanism for concomitant allocational harms. For example, an individual applying for a certain job may be discouraged if all faces returned by Google search on the position do not match their own identity or a recruiter may be influenced towards unfairly prioritizing applicants from the well-represented demographic. We do not explicitly test allocational harms and suggest future research should explore both general and case-specific settings by engaging multiple stakeholders and affected communities \cite{Weidinger2021}. 

\paragraph{Sole focus of bias in face images.} Face datasets were used in original research on implicit bias \cite{Greenwald1998} and have been adopted widely for bias in machine learning contexts, especially in the computer vision community. This motivated our use of face datasets in the subdomain of VL. Note that many well-known large face image datasets present privacy and representational issues, and that FairFace \cite{Karkkainen2021} thus serves an important role in ethical bias research due to its synthetic nature. However, focusing only on face datasets encodes only a narrow presentation of social bias. In reality, social, cultural and historical biases extend far beyond face images, and includes associations on cultural artifacts, practices and geographic localities. We encourage future work on broader presentations of bias and harms in addition to those captured from captioning face datasets.

\paragraph{Code of ethics.} Our method can be applied to reduce representational harm in search queries. Our methods avoid using costly and environmentally-damaging training procedures. We use the privacy-preserving dataset FairFace which avoids potential unconsensual use of face images, but UTKFace may entail privacy risks. We do not employ human annotators in any capacity.

\section{Conclusion}
This paper establishes a framework for measuring and mitigating bias in VL models. Firstly, we demonstrate that ranking metrics (specifically \textit{MaxSkew} and \textit{NDKL}) are effective bias measures. We report these metrics for a range of pretrained VL models for gender and racial bias in photos of faces. Our results confirm previous findings in other domains that (i) more pretraining data correlates with lower model bias, and (ii) training models with SSL can reduce bias. Secondly, we demonstrate a supervised adversarial debiasing method of VL models via learned ``debiasing'' tokens on publicly-available face image datasets with attribute labels.  The proposed method demonstrates a substantial reduction over a suite of bias metrics for gender and race attributes, with feature degradation being wholly mitigable using joint training with an ITC loss on small publicly-available image datasets. 

Future work could include (i) debiasing during the pretraining stage, with SSL showing a promising avenue in that regard, or (ii) defining a wider diversity of attributes such as removing the harmful assumption of binary gender or considering intersectional biases. We encourage researchers in VL to continue to investigate bias in their models, be transparent in documenting model weaknesses using metrics like those proposed in this paper, and seek to apply relatively cheap and easy debiasing protocols like ours. 

Our code, models and debiasing tokens are publicly-available\footnote{See \href{https://github.com/oxai/debias-vision-lang}{https://github.com/oxai/debias-vision-lang}.} for the community to use in the hope that progress can be made towards the safer and fairer use of this technology in society.

\section{Acknowledgements}
The authors would like to thank Mohamed Baioumy, Vit Ruzicka, Jakob Mökander and Laura Weidinger for their helpful feedback. We also thank our anonymous reviewers for their contributions. This work has been supported by the Oxford Artificial Intelligence student society, the EPSRC Centre for Doctoral Training in Autonomous Intelligent Machines \& Systems [EP/S024050/1] (Y.B., A.S.), and the Economic and Social Research Council Grant for Digital Social Science [ES/P000649/1] (H.R.K.).

\bibliography{Bibtex_13022022}
\clearpage
\appendix
\section{Word Embedding Association Test (WEAT)}\label{appendix:weat_math}
The Word Embedding Association Test \cite{Caliskan2017} measures the differential association between a set of two target concepts $\mathcal{C}=\{C_1, C_2\}$ (e.g., `career' and `family') and a set of attributes $\mathcal{A}=\{A_1, ..., A_l\}$ (e.g., `male' and `female'). Here each concept $C_i$ and attribute $A_i$ contain embeddings in a common space for stimuli associated with them (e.g., `office', and `business' for the concept `career', and `boy', `father' and `man' for the attribute `male'). Now the differential association between concepts $C_1$ and  $C_2$ and attributes $A_1$ and $A_2$ is defined as

\begin{dmath}
s(C_1,C_2,A_1,A_2) = \sum_{c_1 \in C_1}{s(c_1, A_1, A_2)} - \sum_{c_2 \in C_2}{s(c_2, A_1, A_2)} ,
\end{dmath}
where, with $\mu$ denoting the arithmetic mean,
\begin{dmath}
s(w, A_1, A_2) = \mu_{a_1 \in A_1}\cos(w, a_1) - \mu_{a_2 \in A_2}\cos(w, a_2)
\end{dmath}
measures the differential association of $w$ with the attributes using cosine similarity. The significance of this association is computed using a permutation test. Denoting all the equal-size partitions of $C_1 \cup C_2$ by $\{(C_1^{i}, C_2^{i})\}^{i}$, we generate a null-hypothesis of no bias and compute the $p$-value
\begin{dmath}
P_{r_i}[s(C_1^{i}, C_2^{i}, A_1, A_2) > s(C_1, C_2, A_1, A_2)]
\end{dmath}

Finally, the effect size, i.e., the normalized measure of the separation between the associations of the targets and attributes, \cite{Caliskan2017} is defined as
\begin{dmath}
\frac{\mu_{c_1 \in C_1}s(c_1, A_1, A_2) - \mu_{c_2 \in C_2}s(c_2, A_1, A_2)}{{\sigma_{c \in C_1\cup C_2}s(c, A_1, A_2)}}
\end{dmath}

In the case of \textit{WEAT}, all attributes and categories are word embeddings. In our experiments, we have cross-modal interactions where the target concepts $\mathcal{C}$ are inferred from the text queries $\mathcal{T}$ and are the corresponding embeddings from the text encoder of the vision-language model, and attributes $\mathcal{A}$ are the image embeddings from the vision encoder.

\section{Ranking metrics}\label{appendix:ranking_metrics}

The following outlines the mathematical implementation of three bias metrics. Let $\tau_y$ be a ranked list of images $\mathcal{I}$ according to their similarity to a text query $T$, and $\tau_T^k$ be the top $k$ images of the list. 

\paragraph{$\pmb{Skew@k}$} measures the difference between the desired proportion of image attributes in $\tau_T^k$ and the actual proportion~\cite{Geyik2019}. For example, given the text query ``this person has a degree in mathematics'', a desired distribution of the image attribute gender could be 50\% to ensure statistical parity.
Let the desired proportion of images with attribute label $A$ in the ranked list be ${{p_{d, T, A}}} \in [0,1]$, and the actual proportion be ${{p_{\tau_T, T, A}}} \in [0,1]$. The resulting $Skew$ of $\tau_{T}$ for an attribute label $A\in \mathcal{A}$ is

\begin{dmath}
Skew_{A}@k(\tau_{T}) = \ln\frac{{{p_{\tau_T, T, A}}}}{{{p_{d, T, A}}}}
\end{dmath}

This measurement gives an indication of possible representational bias \cite{Weidinger2021}, with certain attributes being under-represented in the top $k$ search results (i.e., a negative $Skew_{A_i}@k$). However, $Skew_{A_i}@k$ has a couple of disadvantages: (i) it only measures bias with respect to a single attribute at a time, and so must be aggregated to give a holistic view of the bias over all attributes $A$, and (ii) different chosen values of $k$ gives different results, so more than a single $Skew$ value would need to be computed for each attribute. These disadvantages form the basis of the next two measures, proposed by~\citet{Geyik2019}, which address each of these limitations.

\paragraph{$\pmb{MaxSkew@k}$} is the maximum \textit{Skew@k} among all attribute labels $A$ of the images for a given text query $T$
\begin{dmath}
MaxSkew_@k(\tau_{T}) = \max_{A_i \in \mathcal{A}}Skew_{A_i}@k(\tau_{T}) 
\end{dmath}

This signifies the ``\emph{largest unfair advantage}'' \cite{Geyik2019} belonging to images within a given attribute. The desired outcome is $0$, implying that the real distribution is equal to the desired distribution (e.g., all genders are equally represented in the ranked images, when the desired distribution is uniform).

\paragraph{Normalized Discounted Cumulative KL-Divergence (\textit{NDKL})} employs a ranking bias measure based on the Kullback-Leibler divergence, measuring how much one distribution differs from another. This measure is non-negative, with larger values indicating a greater divergence between the desired and actual distributions of attribute labels for a given $T$. Let $D_{\tau_T^i}$ and $D_T$ denote the discrete distribution of image attributes in $\tau_T^i$ and the desired distribution, respectively. \textit{NDKL} is defined by

\begin{dmath}
\textit{NDKL}(\tau_{T}) = \frac{1}{Z}\sum_{i=1}^{|\tau_{y}|}\frac{1}{\log_{2}(i+1)}d_{KL}(D_{\tau_{T}^{i}}||D_{T})
\end{dmath}

where $d_{KL}(D_{1}||D_{2}) = \sum{_j}D_{1}\ln\frac{D_{1}(j)}{D_{2}(j)}$ is the KL-divergence of distribution $D_{1}$ with respect to distribution $D_{2}$, and $Z = \sum_{i=1}^{|\tau_r|} \frac{1}{\log_2(i+1)}$ is a normalization factor.
The $KL$-divergence of the top-$k$ distribution and the desired distribution is a weighted average of \textit{Skew$_A@k$} measurements (averaging over $A\in \mathcal{A}$). Thus, this aggregation overcomes the first disadvantage of \textit{Skew}, however, \textit{NDKL} is non-negative, and so it cannot distinguish between two ``opposite-biased'' search procedures.

\section{Measuring bias across different model architectures, datasets, and syntactic changes.}\label{appendix:weat_is_weird}
\begin{figure*}[ht]
\centering
    \includegraphics[width=\textwidth]{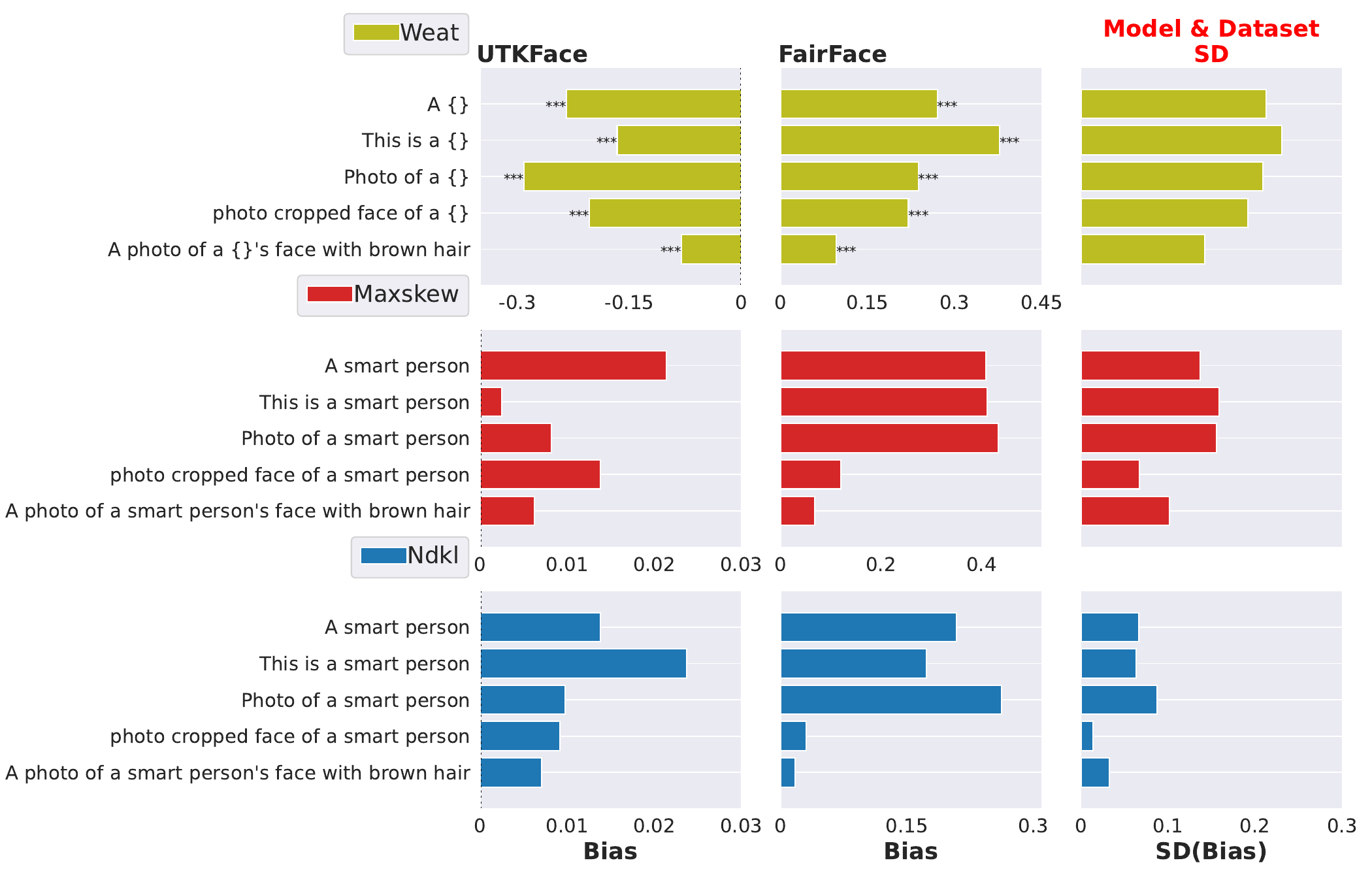}
    \footnotesize
    \captionof{figure}{Bias measures across different combinations of minor syntactic changes, models (RN50, ViT$_{B/16}$, ViT$_{B/32}$), and datasets (FairFace validation set and UTKFace). Bias is measured for gender, and we use the \textit{WEAT} pairwise adjectives concept sets from \citet{Caliskan2017}. Standard deviation of bias measurement is taken over all combinations of model architecture and datasets, for other results we use ViT$_{B/32}$.
    }
    \label{fig:compare_metrics}
\end{figure*}

In \cref{fig:compare_metrics} we report the defined bias measures (\textit{WEAT}, \textit{NDKL} and \textit{MaxSkew}) across changes in vision-language model encoders, datasets and minor syntactic changes to the text queries $T$.

Since \textit{WEAT} uses a template to fill in with concepts, it is not directly comparable to the text queries used in \textit{NDKL} and \textit{MaxSkew}. We report these results only to illustrate the high variance of bias measurement results over small changes in the syntax of templates, model architecture and dataset.

We note that \textit{WEAT} measured on UTKFace has an opposing sign to \textit{WEAT} measured on FairFace. Furthermore, with small syntactic changes in template, \textit{WEAT} produced both positive and negative results on both FairFace and UTKFace. 
This may be explained by the fact that \textit{WEAT} was primarily designed for single word embeddings, while we are using long prompts.~\citet{May2019} found \textit{SEAT} (Sentence Embedding Association Test) to fail for analogous reasons. Accordingly, we implement \textit{MaxSkew@1000} and \textit{NDKL} which show consistent performance in measuring bias across different model architectures, datasets and minor syntactic changes.

\section{Performance effects of learnable text token initialization}\label{appendix:token_init_comparison}
In~\cref{tab:token_eff} we show the effects on zero-shot performance when adding zero-initialized text tokens to the text queries, before any debiasing training has occurred. We note there is a substantial drop in performance in both Flickr image retrieval and CIFAR image classification, with the drop increasing with the number of tokens added in both the prepending and appending settings. This suggests that the reduced ZS performance of the debiased model is not due to the adversarial learning but rather the learnable text tokens which shift the distribution of the text query. 

\begin{table}[]
\centering
\setlength{\tabcolsep}{8pt}
\caption{\small{Results showing effect of prepending or appending with zero-pad initialized text tokens on zero-shot text-to-image retrieval and image classification.}}
\footnotesize
\begin{tabular}{@{}ccrr@{}}
\toprule
\textbf{Token Pos.}     & \textbf{\#tokens} & \multicolumn{1}{c}{{flickr$_{R@5}$}} & \multicolumn{1}{c}{{CIFAR$_{acc}$}} \\ \midrule
\multirow{4}{*}{Prepend} & 0                 & 85.9                                        & 66.5                                       \\
                         & 1                 & 78.3                                        & 57.5                                       \\
                         & 2                 & 70.1                                        & 59.4                                       \\
                         & 3                 & 64.5                                        & 58.5                                       \\ \midrule
\multirow{4}{*}{Append}  & 0                 & 85.9                                        & 66.5                                       \\
                         & 1                 & 68.6                                        & 56.9                                       \\
                         & 2                 & 68.7                                        & 58.5                                       \\
                         & 3                 & 57.0                                        & 54.7                                       \\ \bottomrule
\end{tabular}
\label{tab:token_eff}
\end{table}

\section{Debiasing}
\noindent\textbf{Prepending learnable text tokens.}
We initialize these learnable tokens as the zero-pad embeddings, minimize deviation from the original text embedding to the original text query, and optimize over the learnable tokens -- the rest of the model weights are frozen. However, even with zero-pad initialized token embeddings, token embeddings of prompts are different to their non-prepended counterparts, and so the text-encoder outputs are slightly modified. This results in a degradation of model performance before any training has occurred.

\section{Experimental protocol}

\noindent\textbf{Debiasing implementation.}\label{appendix:debiasing_impl_details}

Models are trained using a NVIDIA GTX Titan X with a batch size of 256. The adversarial classifier is a multilayer perceptron (MLP) with ReLU activation, two hidden layers of size 32, input size equal to the number of training text prompts, and output size equal to the number of sensitive attributes that we debias over, $\text{dim}(A)$. We train with the \textit{Adam} optimizer \cite{Kingma2015} and use learning rates of $2\cdot 10^{-5}$ and $2\cdot 10^{-4}$ for CLIP and the adversarial classifier, respectively. Following an initial two epochs of only training the adversarial model, the CLIP and adversarial model are alternately trained for 10 batches each. Minimal parameter tuning is employed due to the computational costs. Early stopping is implemented if the CLIP model performance as tested on CIFAR100~\cite{Krizhevsky2009}\footnote{Chosen over \textit{IN1K$_{acc}$} monitoring due to its smaller scale.} or Flickr-1k~\cite{Young2014} drops below 50\% of the original accuracy. 
The small size (measured in number or size of hidden layers, or total \# of parameters) of the adversarial model is motivated by the size of its input (fewer than $20$ training prompts) and the size of its output (fewer than $10$ sensitive attributes). We expect even the small adversarial model to remove any linear and reasonable non-linear relationships between the output logits of our vision-language models, i.e., be able to find bias if and when it exists.
For finetuning, we choose to train all combinations of the last three layers of the text encoder (transformer-based with 12 layers total), the last three image encoder layers (also transformer-based with 12 layers) and the two projections from text and image feature space to the embedding space. We purposefully do not choose to train the entire model, as the expected feature quality loss is large, as well as the memory and computational requirements being significantly higher than for training only 25\% of the model's parameters.
We experimented with other implementations of prompt learning than prepending tokens (e.g. appending or adding learned embeddings, and different initializations, e.g. zero-pad, embedding of common token from training corpus, and uniformly random), but these variations showed different feature and bias metric results only at start of training, and no significant change in results. As the number of learned tokens impacted feature quality, we chose 2 tokens as a reasonable trade-off (more tokens giving lower feature quality).
For ITC joint training we used $\lambda = 0.05$ with image-text batches from the Flickr30K training set, unless otherwise specified.

\begin{table*}[!t]
\centering
\footnotesize
\caption{\small{\textbf{Harmful misclassification rate} of FairFace validation images into criminal and non-human categories, by FairFace ethnicity group. We compare between the CLIP Audit paper~\cite{Agarwal2021audit}, a baseline CLIP model, and a CLIP model with debiasing trained on FairFace gender attributes using learned prompt token embeddings.}}
\label{tab:clip_audit}
\setlength{\tabcolsep}{1.5pt}
\begin{tabular}{cccccccccc} 
\toprule
\textbf{Category} & \textbf{Model} & \textbf{Debiased} &\textbf{Black} &	\textbf{White} &	\textbf{Indian} &	\textbf{Latino} &	\begin{tabular}{@{}c@{}}\textbf{Middle} \\ \textbf{Eastern} \end{tabular} &	\begin{tabular}{@{}c@{}}\textbf{Southeast} \\ \textbf{Asian} \end{tabular} &	\begin{tabular}{@{}c@{}}\textbf{East} \\ \textbf{Asian} \end{tabular} \\
\midrule

\multirow{3}{*}{Crime-related} 
    & CLIP Audit \cite{Agarwal2021audit} &\xmark &	16.4 &	24.9 &	24.4 &	10.8 &	19.7 &	4.4	& 1.3 \\
    & CLIP ViT$_{B/16}$ &\xmark & 3.0 &	26.9&	2.7	&4.8	&8.8	&0.5	&0.5\\
    & CLIP ViT$_{B/16}$ & \cmark & 1.7 &	14.9 & 0.1 &	1.7 &	4.5 &	0.4 &	0.3\\ 
\midrule
\multirow{3}{*}{Non-human} & CLIP Audit \cite{Agarwal2021audit} & \xmark &	14.4 &	5.5	& 7.6&	3.7 &	2.0 &	1.9 &	0\\
    &CLIP ViT$_{B/16}$ & \xmark & 0.2	&0.2	& 0.0 &	0.0 &	0.1 &	0.0 &	0.1\\
    & CLIP ViT$_{B/16}$ & \cmark & 0.8 &	0.8	& 0.0 &	0.1	& 0.5& 	0.0	& 0.1\\

\bottomrule
\end{tabular}
\end{table*}
\begin{table*}%
\centering 
\footnotesize 
\caption{\small{\textbf{Comparison of adaptation techniques for debiasing gender} on FairFace via adversarial learning. Bias and zero-shot downstream performance measures are displayed as absolute values with percentage change relative to the pretrained baseline, a CLIP model with ViT$_{B/16}$ architecture.}}
\label{tab:debiasing_comparison}
\begin{tabular}{cllll} 
\toprule
\textbf{Debias}      & \multicolumn{2}{c}{\textbf{Bias Measures~$\downarrow$}} & \multicolumn{2}{c}{\textbf{ZS Performance~$\uparrow$}}  \\ 
\cline{2-5}
\textbf{~Adaptation} & \textit{MaxSkew@1000} & \textit{NDKL} & {flickr$_{R@5}$} & {CIFAR$_{acc}$}                 \\ 
\hline
PT baseline                    & 0.233        & 0.103                           & 86.1           & 66.5                          \\ 
\hline
\textbf{Prompt}             & \textbf{0.073\scriptsize{(-69\%)}} & \textbf{0.021\scriptsize{(-80\%)}}     & 64.2\scriptsize{(-25\%)}         & \textbf{54.3\scriptsize{(-18\%)}}                        \\ 

Proj. layer          & 0.642\scriptsize{(+176\%)}      & 0.561\scriptsize{(+443\%)}        & 62.3\scriptsize{(-28\%)}          &40.6\scriptsize{(-39\%)}                        \\
Text encoder         & 0.691\scriptsize{(+197\%)}      & 0.688\scriptsize{(+566\%)}     & \textbf{67.\scriptsize{8(-21\%)}}          &43.4\scriptsize{(-35\%)}                        \\
Full finetuning     & 0.688\scriptsize{(+195\%)}     & 0.664\scriptsize{(+543\%)}       & 18.6\scriptsize{(-78\%)}          &6.6\scriptsize{(-90\%)}                      \\
\bottomrule
\end{tabular}

\end{table*}
\section{Harmful Zero-Shot Misclassification}\label{appendix:clip_audit_impl}
We follow the protocol of \citet{Agarwal2021audit} by using CLIP to classify images from the FairFace validation set into different categories, the $7\cdot2=14$ FairFace ethnicity-gender class pairs, non-human categories (animal, gorilla, chimpanzee, and orangutan) and crime-related words (thief, criminal and suspicious person). We then look at the percentage of images that are misclassified into the non-human and crime classes. The original implementation is lacking in details, and it is unclear if they use a template approach. We use the template ``a photo of a \{\}", since it is the standard for all other CLIP measurements. We also tried performing the test without using a query template but classification accuracy was significantly reduced for all images. \\ \indent ~\cref{tab:clip_audit} shows the results directly taken from \citet{Agarwal2021audit} alongside results from our implementation with the pretrained baseline CLIP ViT$_{B/16}$.
Our gender-debiased model trained on FairFace has a lower misclassification rate into crime-related classes than the pretrained baseline. While the non-human misclassification rate was marginally higher than baseline, the absolute rates are still comparable and very low ($<$1\%). For all ethnicities with misclassification rates greater than 1\% from the pretrained baseline, our debiased model reduces the rate by half or more (-43\% to -96\%).

\section{Additional Results}\label{appendix:debiasing_comparison}

In ~\cref{tab:debiasing_comparison} we show the result of finetuning over different parts of the model as well as pure prompt learning, all with pure adversarial training. The strong regularization from having few learned embeddings keeps the feature quality at an acceptable level, and finetuning larger parts of the model lowered model performance to an unacceptable level very quickly during training.

\end{document}